%% file: 2017_EvoNet2_arxiv.tex
\documentclass[10pt,twocolumn,letterpaper]{article}

\usepackage{iccv}
\usepackage{times}
\usepackage{epsfig}
\usepackage{graphicx}
\usepackage{amsmath}
\usepackage{amssymb}
\usepackage{multirow}
\usepackage{float}
\usepackage{subcaption}

\usepackage[pagebackref=true,breaklinks=true,letterpaper=true,colorlinks,bookmarks=false]{hyperref}

\iccvfinalcopy % *** Uncomment this line for the final submission

\def\iccvPaperID{42} % *** Enter the ICCV Paper ID here

\ificcvfinal\fi
\begin{document}

%%%%%%%%% TITLE
\title{The Mating Rituals of Deep Neural Networks: Learning Compact Feature Representations through Sexual Evolutionary Synthesis}

\author{Audrey G. Chung \and Mohammad Javad Shafiee \and Paul Fieguth \and Alexander Wong\\
University of Waterloo\\
200 University Ave W, Waterloo, ON, N2L 3G1\\
{\tt\small agchung@uwaterloo.ca}
%\and
}

\maketitle
%\thispagestyle{empty}

%%%%%%%%% ABSTRACT
\begin{abstract}
\input{./content/Abstract.tex}
\end{abstract}

%%%%%%%%% BODY TEXT
\section{Introduction}
\label{Introduction}
\input{./content/Introduction.tex}

\section{Methods}
\label{Methods}
\input{./content/Methods.tex}

\section{Results}
\label{Results}
\input{./content/Results.tex}

\section{Conclusion}
\label{Conclusion}
\input{./content/Conclusion.tex}

%\section*{Acknowledgement}
%This work was supported by the Natural Sciences and Engineering Research Council of Canada, Ontario Ministry of Economic Development and Innovation  and Canada Research Chairs Program. The authors also thank Nvidia for the GPU hardware used in this study through the Nvidia Hardware Grant Program.

{\small
\bibliographystyle{ieee}
\bibliography{EvoNet2}
}

\end{document}

%% file: content/Abstract.tex
\textit{Evolutionary deep intelligence} was recently proposed as a method for achieving highly efficient deep neural network architectures over successive generations. Drawing inspiration from nature, we propose the incorporation of sexual evolutionary synthesis. Rather than the current asexual synthesis of networks, we aim to produce more compact feature representations by synthesizing more diverse and generalizable offspring networks in subsequent generations via the combination of two parent networks. Experimental results were obtained using the MNIST and CIFAR-10 datasets, and showed improved architectural efficiency and comparable testing accuracy relative to the baseline asexual evolutionary neural networks. In particular, the network synthesized via sexual evolutionary synthesis for MNIST had approximately double the architectural efficiency (cluster efficiency of 34.29$\times$ and synaptic efficiency of 258.37$\times$) in comparison to the network synthesized via asexual evolutionary synthesis, with both networks achieving a testing accuracy of $\sim$97\%.

%Evolutionary deep intelligence was recently proposed as a method for achieving highly efficient deep neural network architectures over successive generations. Inspired by nature, we propose the incorporation of sexual evolutionary synthesis. Rather than the current asexual synthesis of networks, we aim to produce more compact feature representations by synthesizing more diverse and generalizable offspring networks in subsequent generations via the combination of two parent networks. Experimental results were obtained using the MNIST and CIFAR-10 datasets, and showed improved architectural efficiency and comparable testing accuracy relative to the baseline asexual evolutionary neural networks. In particular, the network synthesized via sexual evolutionary synthesis for MNIST had double the architectural efficiency (cluster efficiency of 34.29X and synaptic efficiency of 258.37X) in comparison to asexual evolutionary synthesis, with both networks achieving a testing accuracy of ~97%.

%, and SVHN
%we propose the incorporation of importance-weighted mating during the evolutionary synthesis stage. Rather than an arbitrary combination of parent networks, we aim synthesize stronger offspring networks in subsequent generations by specifically passing on beneficial traits from the parent networks. Experimental results were obtained using the CIFAR-10 and STL-10 %, MNIST, 
%object classification datasets, and show improved performance relative to the baseline asexual evolutionary neural network. 

%% file: content/Introduction.tex
Deep learning methods, especially deep neural networks~\cite{LeCun2015, Bengio2009, Graves2013, Tompson2014}, have recently exploded in popularity due to their demonstrated ability to significantly improve the performance over other machine learning methods in various challenging areas of research. However, this boost in performance of deep neural networks is largely attributed to increasingly large model sizes, resulting in growing storage and memory requirements. 

These computational requirements make high-performance deep neural networks infeasible for devices without access to cloud computing. For many practical situations such as self-driving cars and smartphone applications, the available computing resources are limited to low-power, embedded GPUs and CPUs; with such limited computational power and storage, smaller and more compact versions of deep neural networks are highly desirable. As such, research into compact feature representations via highly efficient deep neural networks has been conducted, and methods have been developed for significantly reducing the memory and computational requirements with minimal drop in performance.

\input{./content/Background.tex}

Rather than attempting to compress existing deep neural networks into smaller and more compact representations directly, Shafiee~\textit{et al.}~\cite{Shafiee2016} proposed an entirely novel concept: \textit{Can deep neural networks naturally evolve to be highly efficient?} Inspired by biological evolution, Shafiee~\textit{et al.} developed an \textit{evolutionary deep intelligence} approach to produce highly efficient and compact deep neural networks by allowing these networks to synthesize new networks with increasingly compact representations and naturally sparsify over successive generations. Biological evolutionary mechanisms are mimicked via three computational constructs: i) heredity, ii) natural selection, and iii) random mutation.

While previous studies~\cite{Angeline1994, Stanley2002, Stanley2005, Gauci2007, Tirumala2016} have been conducted that leverage the idea of using evolutionary techniques to generate and train neural networks, there are key differences between these and the evolutionary deep intelligence method proposed by Shafiee~\textit{et al.}~\cite{Shafiee2016}. Past works have primarily focused on improving a network's training and accuracy, while evolutionary deep intelligence shifts the focus to organically synthesizing networks with high architectural efficiency. In addition, these previous studies use classical evolutionary computation approaches such as genetic algorithms and evolutionary programming, while Shafiee~\textit{et al.} introduced a novel probabilistic framework that models genetic encoding and environmental conditions via probability distributions.

More recently, Shafiee \textit{et al.} proposed a modification of the original evolutionary deep intelligence approach via synaptic cluster-driven genetic encoding~\cite{Shafiee2016_2}. Further investigating the genetic encoding scheme used to mimic heredity, Shafiee \textit{et al.} proposed the incorporation of synaptic clustering into the genetic encoding scheme, and introduced a multi-factor synapse probability model. Modelling the synaptic probability as a product of the probability of synthesis of a particular cluster of synapses and the probability of synthesis of a particular synapse within the cluster, this new genetic encoding scheme demonstrated state-of-the-art performance while producing significantly more efficient network architectures and compact feature representations specifically tailored for GPU-accelerated applications.

The current work in evolutionary deep intelligence~\cite{Shafiee2016,Shafiee2016_2}, however, formulates the evolutionary synthesis process based on asexual reproduction; that is, offspring neural networks are synthesized by stochastically sparsifying a clone of their parent network. While effective at synthesizing efficient networks with comparable testing accuracies, asexual evolutionary synthesis results in a limited range of possible offspring networks as the offspring network structure is highly constrained by the parent network. Motivated by the aim of promoting diversity in evolutionary deep neural networks, we explore the use of sexual reproduction when synthesizing offspring network architectures in this study.

Evolutionarily speaking, sexual reproduction is thought to have developed in living organisms due to the fact that it favours the survival of groups rather than individuals by allowing for accelerated adaptation to changing environments via the combination of mutations occurring in distinct individuals in a single descendant~\cite{Fisher1930,Muller1932}. Relative to asexual reproduction, sexual reproduction has the potential to accelerate evolution by several orders of magnitude~\cite{Crow1965}, with its effects most prominent in a large population with a high frequency of beneficial mutations. This motivates the idea that the use of sexual reproduction in evolutionary synthesis can accelerate the generation-by-generation development of useful compact feature representations.

To evaluate the validity of sexual reproduction in evolutionary deep intelligence, we propose an extension of Shafiee \textit{et al.}'s cluster-driven genetic encoding approach~\cite{Shafiee2016_2} via the incorporation of a second parent network during the synthesis of an offspring network at each generation. The methodology for the proposed model is described in Section~\ref{Methods}. The experimental setup and results are presented in Section~\ref{Results}. Lastly, conclusions and future works are discussed in Section~\ref{Conclusion}.

%% file: content/Background.tex
One of the first approaches for adapting the size of a neural network was optimal brain damage~\cite{LeCun1989}. The method removed unimportant weights (as determined using the second derivative of the objective function as a saliency approximation of a parameter) from the network to improve network generalizability, increase the speed of learning, and reduce the number of training sampled required.

Gong \textit{et al.}~\cite{Gong2014} proposed a network compression framework where vector quantization was leveraged to shrink the storage requirements of deep neural networks trained for computer vision tasks. Gong \textit{et al.} noted that vector quantization has clear advantages over existing matrix factorization methods, and found a good balance between model size and accuracy could be achieved via the application of \mbox{k-means}	 clustering to the weights or product quantization.

Han \textit{et al.}~\cite{Han2015} introduced deep compression to address the limitations of computational power and memory that comes with embedded systems via a three stage pipeline: pruning, trained quantization, and Huffman coding. The method reduced the storage requirements of a neural network by 35x to 49x with no loss in accuracy. Han~\textit{et al.}~\cite{Han2015_2} also reduced the storage and computational requirements of neural networks with no drop in accuracy by training a network to learn which weights are important, pruning the unimportant connections, and retraining the network to fine tune the remaining weights.

Another method for deep compression is hashing~\cite{Chen2015}, which uses a low-cost hash function to group network weights into hash buckets with a single shared parameter value. Exploiting the redundancy in both network layers, the HashedNet architecture leverages the idea of weight-sharing and allows for considerable savings in terms of memory and storage.

Other methods for reducing the computational requirements of neural networks include low rank approximations~\cite{Jaderberg2014,Ioannou2015}. Jaderberg~\textit{et al.}~\cite{Jaderberg2014} used low-rank expansions to speed up the computation of deep neural networks (specifically the convolutional layers of convolutional neural networks) by exploiting cross-channel or filter redundancy to construct a low rank basis of filters. Similarly, Ioannou \textit{et al.}~\cite{Ioannou2015} created computationally efficient networks using low rank representations of convolutional filters by learning a set of small basis filters that are then combined into more complex filters.

Sparsity learning~\cite{Feng2015,Liu2015,Wen2016} is another strategy used to sparsify deep neural networks. Feng and Darrell~\cite{Feng2015} demonstrated a novel method for learning components of the structure of a neural network by incorporating the Indian Buffet Process prior; especially effective when there is limited labelled training data, this method captures complex data distributions in an unsupervised generative manner. \mbox{Liu~\textit{et al.}~\cite{Liu2015}} showed how to reduce the number of parameters in neural networks via sparse decomposition by exploiting both intra-channel and inter-channel redundancy. Lastly, Wen~\textit{et al.}~\cite{Wen2016} recently proposed a Structured Sparsity Learning (SSL) method to regularize the structures within deep neural networks (e.g., filters shapes, channels, layer depth).

%% file: content/Methods.tex
\begin{figure*}[t]
	\centering
	\includegraphics[width=\textwidth]{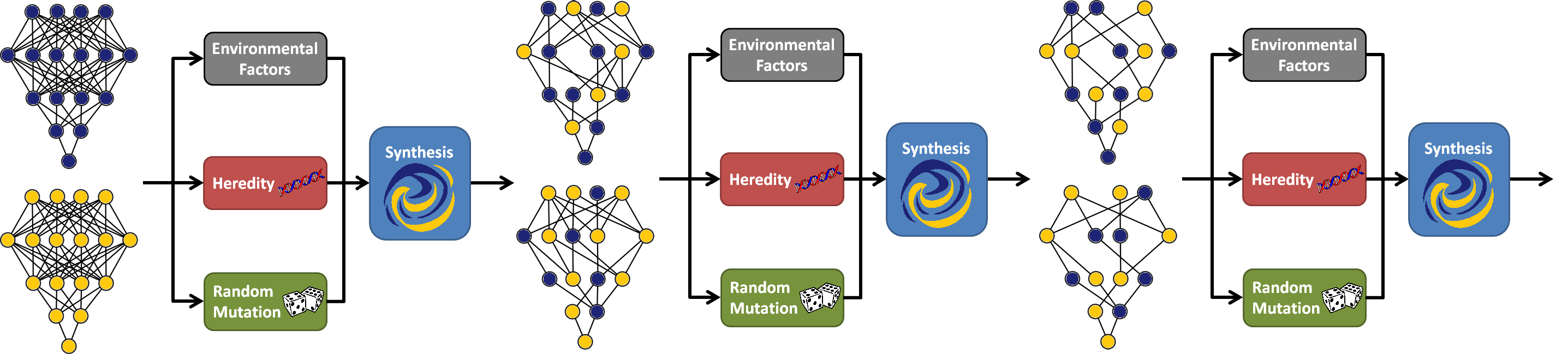}
	\caption{The proposed evolutionary synthesis process over successive generations as an extension of cluster-driven genetic encoding. The effects of sexual evolutionary synthesis are explored via the incorporation of a second parent network during the synthesis of offspring networks. At each generation, two parent networks from the preceding generation are combined via a mating function to synthesize new offspring networks.}
	\label{fig_AlgDesign}
\end{figure*}

In this work, we propose an extension of Shafiee \textit{et al.}'s cluster-driven genetic encoding~\cite{Shafiee2016_2} via an adaptation of the network synthesis process towards more compact feature representations using sexual evolutionary synthesis. To explore the effects of sexual evolutionary synthesis when synthesizing offspring network architectures, we augment the evolutionary deep intelligence scheme in~\cite{Shafiee2016_2} to incorporate a second parent network during the process of synthesizing new offspring networks as shown in Figure~\ref{fig_AlgDesign}. At each generation, two parent networks from the preceding generation are combined via a mating function to synthesize new offspring networks containing information from both parent networks.

\subsection{Sexual Evolutionary Synthesis}
Let the network architecture be formulated as $\mathcal{H}(N,S)$, where $N$ denotes the set of possible neurons and $S$ the set of possible synapses in the network. Each neuron $n_j \in N$ is connected to neuron $n_k \in N$ via a set of synapses $\bar{s} \subset S$, such that the synaptic connectivity $s_j \in S$ has an associated $w_j \in W$ to denote the connection's strength. In the seminal paper on evolutionary deep intelligence~\cite{Shafiee2016}, the synthesis probability $P(\mathcal{H}_{g}|\mathcal{H}_{g-1}, \mathcal{R}_{g})$ of a new network at generation $g$ is approximated by the synaptic probability $P(S_g|W_{g-1}, R_{g})$ to emulate heredity through the generations of networks, and is also conditional on an environmental factor model $\mathcal{R}_g$ to imitate natural selection via a changing environment for successive generations of networks to adapt to. That is, the synthesis probability can be formulated as follows:
\begin{align}
	P(\mathcal{H}_{g}|\mathcal{H}_{g-1}, \mathcal{R}_{g}) \simeq P(S_g | W_{g-1}, R_{g}).
\end{align}

More recently, Shafiee \textit{et al.}~\cite{Shafiee2016_2} reformulated the synthesis probability to incorporate the multi-factor synaptic probability model and different quantitative environmental factor models at the synapse and cluster levels:

\begin{align}
	P(\mathcal{H}_{g}|&\mathcal{H}_{g-1}, \mathcal{R}_g) = \nonumber \\
	& \prod_{c \in C} \Big[ P(s_{g,c}|W_{g-1}, \mathcal{R}_{g}^c) \cdot \prod_{j \in c} P(s_{g,j}|w_{g-1,j}, \mathcal{R}_g^s) \Big]
\end{align}
where $\mathcal{R}_{g}^c$ and $\mathcal{R}_{g}^s$ represent the environmental factor models enforced during the synaptic cluster synthesis and the synapse synthesis, respectively. $P(s_{g,c}|W_{g-1}, \mathcal{R}_{g}^c)$ represents the probability of synthesis for a given cluster of synapses $s_{g,c}$; that is, $P(s_{g,c}|W_{g-1}, \mathcal{R}_{g}^c)$ denotes the likelihood that a synaptic cluster $s_{g,c}$ will exist in the network architecture in generation $g$ given the cluster's synaptic strength in generation $g-1$ and the cluster-level environmental factor model. Comparably, $P(s_{g,j}|w_{g-1,j}, \mathcal{R}_g^s)$ represents the likelihood of the existence of synapse $j$ within the synaptic cluster $c$ in generation $g$ given the synaptic strength in the previous generation $g-1$ and synapse-level environmental factor model. This multi-factor probability model encourages both the persistence of strong synaptic clusters and the persistence of strong synaptic connectivity over successive generations~\cite{Shafiee2016_2}.

With asexual evolutionary synthesis, however, a limited range of possible offspring networks is explored as the structure of each network is constrained by its parent network. Taking inspiration from nature, we aim to increase the diversity and compactness of evolutionary deep neural networks by incorporating information from multiple parent networks when synthesizing a new offspring network. In addition to increasing diversity, previous work in the field of evolutionary biology has concluded that sexual reproduction will accelerate adaptation to a new environment given that the genetic variance arises due to a changing environment~\cite{Smith1968}. This motivates the idea that more efficient and diverse offspring networks with increasingly compact feature representations can be synthesized in fewer generations using sexual evolutionary synthesis, particularly in the case of non-stationary environmental factor models.

In this work, we propose a further modification of the synthesis probability $P(\mathcal{H}_{g}|\mathcal{H}_{g-1}, \mathcal{R}_g)$ via the incorporation of a two-parent synthesis process to drive network diversity and adaptability by mimicking sexual reproduction. Thus far for the $i^{\text{th}}$ synthesized network, the cluster synthesis probability $P(s_{g,c}|W_{g-1}, \mathcal{R}_{g}^c)$ and the synapse synthesis probability $P(s_{g,i}|w_{g-1,i}, \mathcal{R}_{g}^s)$ have been conditional on the network architecture and synaptic strength of a single parent network in the previous generation and the environmental factor models. To explore the effects of sexual evolutionary synthesis in evolutionary deep intelligence, we reformulate the synthesis probability to combine the cluster and synapse probabilities of two parent networks, e.g., $\mathcal{H}_A$ and $\mathcal{H}_B$, during the synthesis of an offspring network via some cluster-level mating function $\mathcal{M}_c(\cdot)$ and some synapse-level mating function $\mathcal{M}_s(\cdot)$:

\begin{align}
P(\mathcal{H}_{g,i}|&\mathcal{H}_{A}, \mathcal{H}_{B}, \mathcal{R}_{g}) =  \nonumber \\
& \prod_{c \in C} \Big[ P(s_{g,c}|\mathcal{M}_c(W_{\mathcal{H}_A},W_{\mathcal{H}_B}), \mathcal{R}_{g(i)}^c) \cdot \nonumber \\
& \prod_{j \in c} P(s_{g,j}|\mathcal{M}_s(w_{\mathcal{H}_A,j},w_{\mathcal{H}_B,j}), \mathcal{R}_{g(i)}^s) \Big].
%	P(\mathcal{H}_g) = \prod_{c \in C} \Big[ \mathcal{F}_c(\mathcal{E}) \cdot \frac{P_A(\bar{s}_{g,c}|W_{g-1}) + P_B(\bar{s}_{g,c}|W_{g-1})}{2} \cdot \nonumber \\
%					   \prod_{i \in c} \mathcal{F}_s(\mathcal{E}) \cdot \frac{P_A(s_{g,i}|w_{g-1,i}) + P_B(s_{g,i}|w_{g-1,i})}{2} \Big]
\end{align} 
%where $P_A(\bar{s}_{g,c}|W_{g-1})$ and $P_B(\bar{s}_{g,c}|W_{g-1})$ are the cluster synthesis probabilities of parent A and B, respectively, and $P_A(s_{g,i}|w_{g-1,i})$ and $P_B(s_{g,i}|w_{g-1,i})$ are the synapse synthesis probabilities of parent A and B, respectively.

\subsection{Mating Rituals of Deep Neural Networks}
In this work, we restrict the the parent networks, $\mathcal{H}_A$ and $\mathcal{H}_B$, to the immediately preceding generation; that is, for an offspring $\mathcal{H}_{g,i}$ at generation $g$, the parent networks $\mathcal{H}_A$ and $\mathcal{H}_B$ are from generation $g-1$. We propose the cluster-level and synapse-level mating functions to be as follows:

\begin{align}
	\mathcal{M}_c(W_{\mathcal{H}_A},W_{\mathcal{H}_B}) &= \alpha_c W_{\mathcal{H}_A} + \beta_c W_{\mathcal{H}_B}\\
	\mathcal{M}_s(w_{\mathcal{H}_A,j},w_{\mathcal{H}_B,j}) &= \alpha_s w_{\mathcal{H}_A,j} + \beta_s w_{\mathcal{H}_B,j}
\end{align}
where $W_{\mathcal{H}_A}$ and $W_{\mathcal{H}_B}$ represent the cluster's synaptic strength for parent networks $\mathcal{H}_A$ and $\mathcal{H}_B$, respectively. Similarly, $w_{\mathcal{H}_A,j}$ and $w_{\mathcal{H}_B,j}$ represent the synaptic strength of a synapse $j$ within cluster $c$ for parent networks $\mathcal{H}_A$ and $\mathcal{H}_B$, respectively.

\subsection{Realization of Genetic Encoding}
In this study, we employ the simple realization of cluster-driven genetic encoding proposed in~\cite{Shafiee2016_2}. The probability of synthesis for a given synapse cluster $s_{g,c}$ is realized as:

\begin{align}
	P(s_{g,c}|W_{g-1}) = \text{exp} \Big( \frac{\Sigma_{i \in c} \lfloor w_{g-1,i} \rfloor}{Z} - 1 \Big)
\end{align}

\noindent where $\lfloor \cdot \rfloor$ encodes a synaptic weight truncation and $Z$ is the normalization factor required to construct a probability distribution, i.e., $P(s_{g,c}|W_{g-1}) \in [0,1]$. The truncation of synaptic weights lessens the impact of weak synapses in a synaptic cluster. 

Similarly, the probability of synthesis for a particular synapse $s_{g,i}$ within a synaptic cluster $c$ is realized as: 

\begin{align}
	P(s_{g,i}|w_{g-1,i}) = \text{exp} \Big( \frac{w_{g-1,i}}{a} - 1 \Big)
\end{align}

\noindent where $z$ is a layer-wise normalization factor. This genetic encoding scheme allows for the simultaneous consideration of both inter-synapse relationships and individual synapse strength~\cite{Shafiee2016_2}.

%% file: content/Results.tex
\subsection{Experimental Setup}
\begin{figure}[t!]
	\centering
	\includegraphics[width=\linewidth]{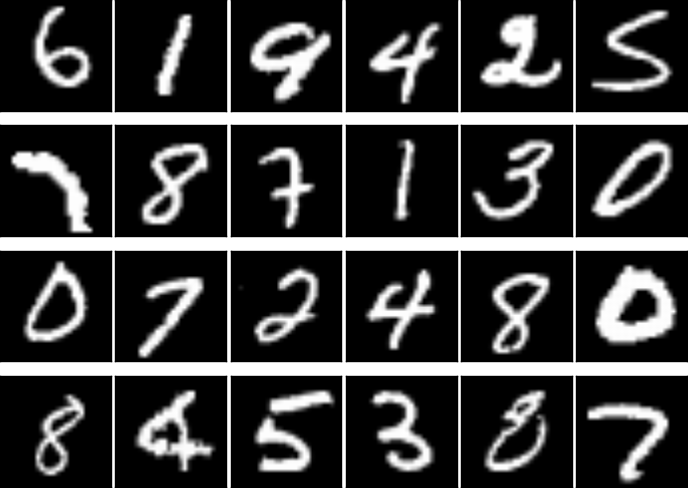}
	\caption{Sample images from the MNIST hand-written digits dataset~\cite{LeCun1998}.}
	\label{fig_MNIST}
\end{figure}
\begin{figure}[t!]
	\centering
	\includegraphics[width=\linewidth]{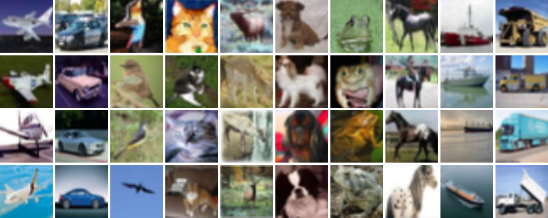}
	\caption{Sample images from the CIFAR-10 object classification dataset~\cite{Krizhevsky2009}.}
	\label{fig_CIFAR}
\end{figure}
%\begin{figure}[t!]
%	\centering
%	\includegraphics[width=\linewidth]{./gfx/SVHN.png}
%	\caption{Sample images from the SVHN house numbers dataset~\cite{Netzer2011}.}
%	\label{fig_SVHN}
%\end{figure}

The asexual and sexual evolutionary synthesis of deep neural networks were performed over several generations, and the effects of sexual evolutionary synthesis relative to asexual evolutionary synthesis were explored using the MNIST~\cite{LeCun1998} hand-written digits and CIFAR-10~\cite{Krizhevsky2009} object classification datasets with the first generation ancestor networks trained using the LeNet-5 architecture~\cite{LeCun1998_LeNet}. Figure~\ref{fig_MNIST} and Figure~\ref{fig_CIFAR} shows sample images from the MNIST and CIFAR-10 datasets, respectively.%STL-10~\cite{Coates2010} and ~\ref{fig_STL-10} 
% , and SVHN~\cite{Netzer2011} house numbers 

Similar to Shafiee \textit{et al.}'s work~\cite{Shafiee2016_2}, we designed the environmental factor models $\mathcal{R}^c_{g(i)}$ and $\mathcal{R}^s_{g(i)}$ to enforce that an offspring deep neural network is limited to $70\%$ of the total number of synapses in its parent network in the previous generation; this allows increasingly more compact feature representations and for the synthesized deep neural networks to become progressively more efficient in the successive generations while minimizing any loss in accuracy. In addition, each filter (i.e., collection of kernels) was considered as a synaptic cluster in the multi-factor synapse probability model, and both the synaptic efficiency and cluster efficiency were assessed along with testing accuracy.

\subsection{Experimental Results}
%\begin{figure*}
%\centering
%\begin{tabular}{ccc}
%\includegraphics[width=.32\textwidth]{gfx/MNIST_TestingAccuracy.png}&
%\includegraphics[width=.32\textwidth]{gfx/MNIST_SynapticEfficiency.png}&
%\includegraphics[width=.32\textwidth]{gfx/MNIST_ClusterEfficiency.png}\\
%\includegraphics[width=.32\textwidth]{gfx/CIFAR_TestingAccuracy.png}&
%\includegraphics[width=.32\textwidth]{gfx/CIFAR_SynapticEfficiency.png}&
%\includegraphics[width=.32\textwidth]{gfx/CIFAR_ClusterEfficiency.png}\\
%\includegraphics[width=.32\textwidth]{gfx/SVHN_TestingAccuracy.png}&
%\includegraphics[width=.32\textwidth]{gfx/SVHN_SynapticEfficiency.png}&
%\includegraphics[width=.32\textwidth]{gfx/SVHN_ClusterEfficiency.png}\\
%\end{tabular}
%\caption{.}
%\label{fig_Results}
%\end{figure*}

%\begin{figure}
%\centering
%\begin{tabular}{p{\linewidth}}
%\includegraphics[width=0.98\linewidth]{gfx/SVHN_TestingAccuracy.png}\\
%(a) SVHN testing accuracy for networks using asexual and sexual evolutionary synthesis. \\
%\includegraphics[width=0.98\linewidth]{gfx/SVHN_SynapticEfficiency.png}\\
%(b) SVHN synaptic efficiency for networks synthesized using asexual and sexual evolutionary synthesis. \\
%\includegraphics[width=0.98\linewidth]{gfx/SVHN_ClusterEfficiency.png}\\
%(c) SVHN cluster efficiency for networks synthesized using asexual and sexual evolutionary synthesis. \\
%\end{tabular}
%\caption{SVHN testing accuracy and network efficiency (synaptic and cluster) for networks synthesized using asexual (red) and sexual (blue) evolutionary synthesis.}
%\label{fig_SVHNResults}
%\end{figure}

\begin{figure}
\centering
\begin{tabular}{p{\linewidth}}
\includegraphics[width=0.98\linewidth]{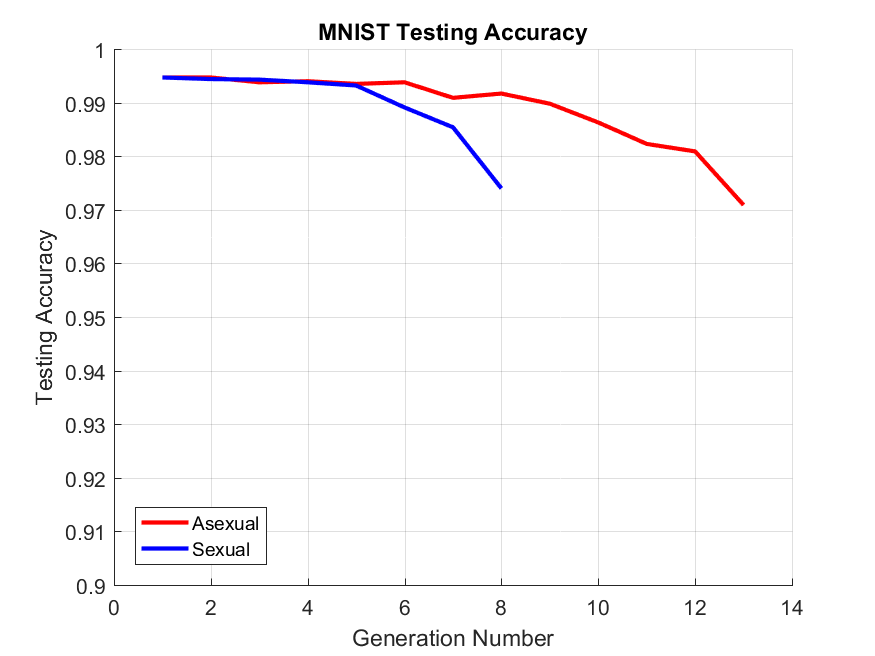}\\
(a) MNIST testing accuracy vs. generations for synthesized networks. \\
\includegraphics[width=0.98\linewidth]{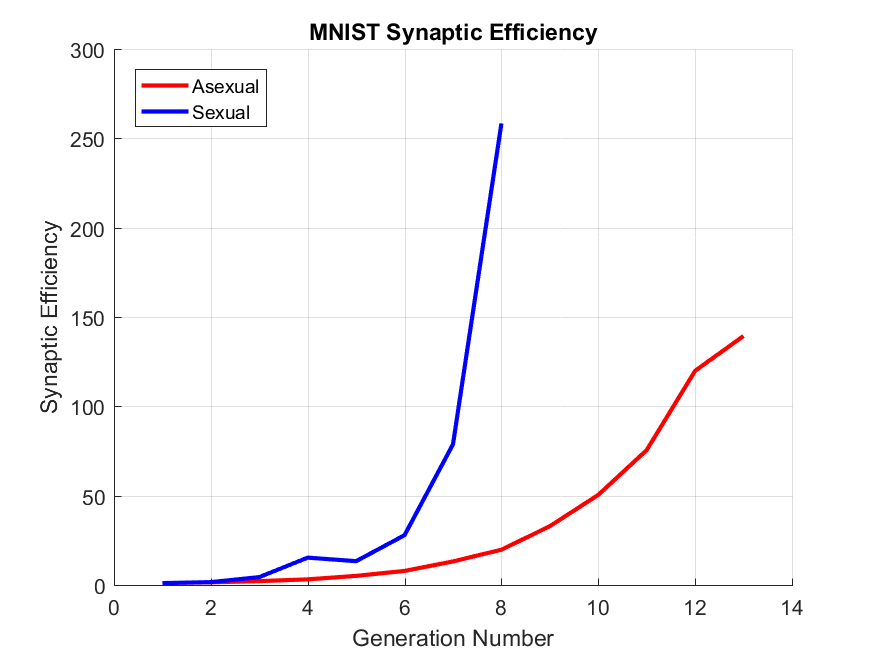}\\
(b) MNIST synaptic efficiency vs. generations for synthesized networks. \\
\includegraphics[width=0.98\linewidth]{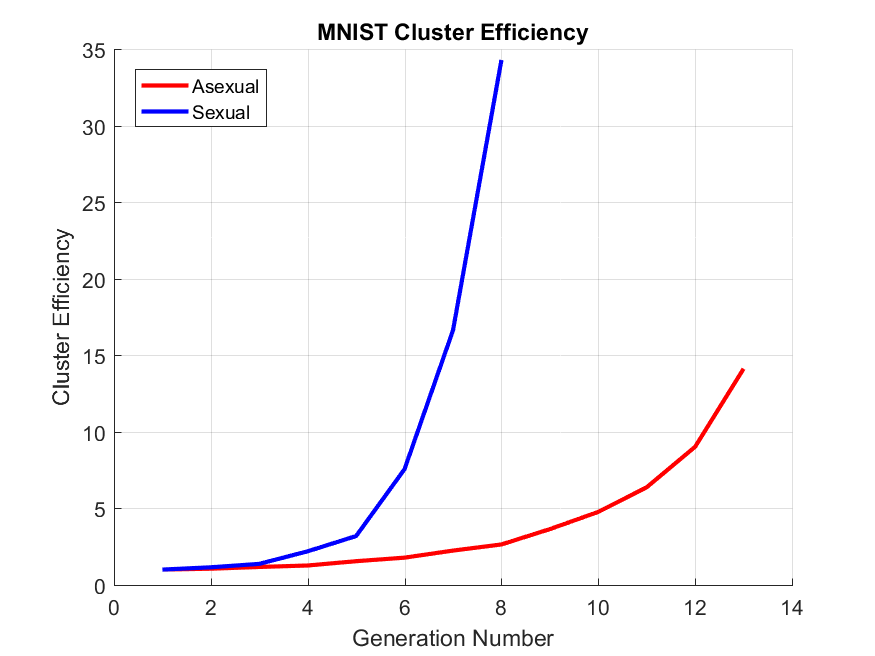}\\
(c) MNIST cluster efficiency vs. generations for synthesized networks. \\
\end{tabular}
\caption{MNIST testing accuracy and network efficiency (synaptic and cluster) for networks synthesized using asexual (red) and sexual (blue) evolutionary synthesis.}
\label{fig_MNISTResults}
\end{figure}

\begin{figure}
\centering
\begin{tabular}{p{\linewidth}}
\includegraphics[width=0.98\linewidth]{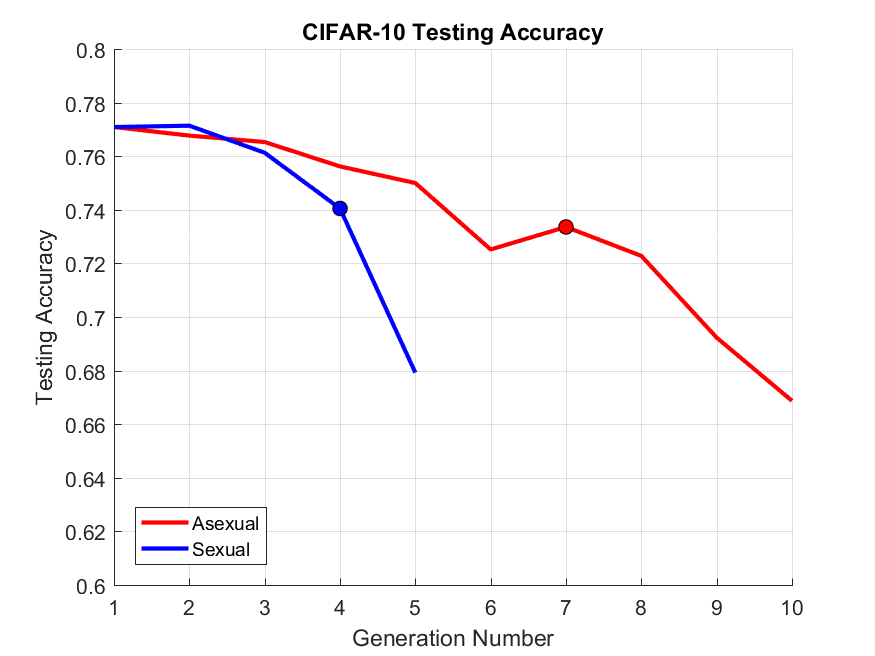}\\
(a) CIFAR-10 testing accuracy vs. generations for synthesized networks; point at 3 -- 4\% accuracy drop marked. \\
\includegraphics[width=0.98\linewidth]{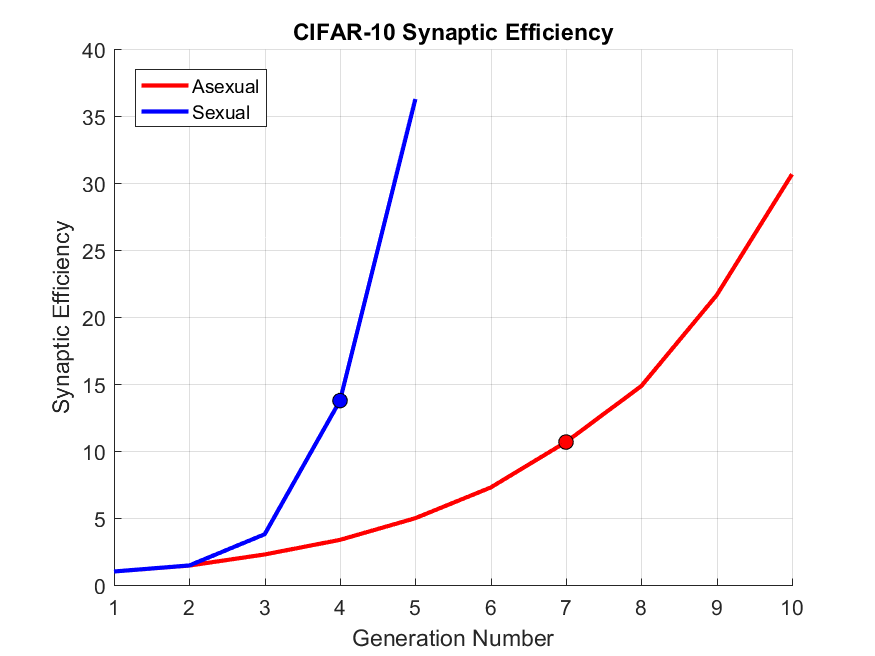}\\
(b) CIFAR-10 synaptic efficiency vs. generations for synthesized networks; point at 3 -- 4\% accuracy drop marked. \\
\includegraphics[width=0.98\linewidth]{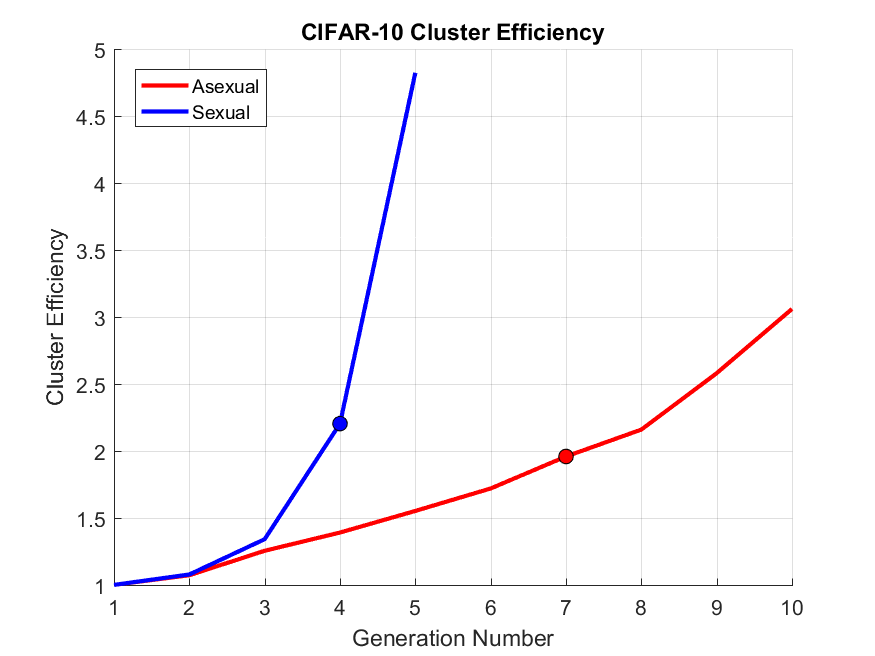}\\
(c) CIFAR-10 cluster efficiency vs. generations for synthesized networks; point at 3 -- 4\% accuracy drop marked. \\
\end{tabular}
\caption{CIFAR-10 testing accuracy and network efficiency (synaptic and cluster) for networks synthesized using asexual (red) and sexual (blue) evolutionary synthesis.}
\label{fig_CIFARResults}
\end{figure}

An extension of Shafiee \textit{et al.}'s cluster-driven genetic encoding scheme~\cite{Shafiee2016_2} for asexual evolutionary synthesis, the use of sexual evolutionary synthesis during the synthesis of offspring networks for evolutionary deep intelligence was explored in this study. At each generation, the network testing accuracy was evaluated and the corresponding architectural efficiency was assessed in terms of cluster efficiency (defined as the reduction in the total number of kernels in a network relative to the first generation ancestor network) and synaptic efficiency (defined as the reduction in the total number of synapses in a network relative to the first generation ancestor network). Figure~\ref{fig_MNISTResults} and Figure~\ref{fig_CIFARResults} show the testing accuracy, synaptic sparsity, and cluster sparsity of networks synthesized using asexual and sexual evolutionary synthesis as a function of generation number, and evaluated on the MNIST and CIFAR-10 datasets, respectively. Note that in both the asexual and sexual evolutionary synthesis cases, there is a trade-off between testing accuracy and architectural efficiency, i.e., testing accuracy decreases as synaptic efficiency and cluster efficiency increase.

Figure~\ref{fig_MNISTResults} shows the MNIST testing accuracy and network efficiency (synaptic and cluster) for networks synthesized using asexual (red) and sexual (blue) evolutionary synthesis. Figure~\ref{fig_MNISTResults} (a) shows the testing accuracy for networks synthesized using asexual and sexual evolutionary synthesis. For this experiment, the original fully-trained ancestor network (generation 1) had a testing accuracy of $99.47\%$ with 143,136 synapses and 7,200 kernels (corresponding to a 1-channel input LeNet architecture~\cite{LeCun1998_LeNet}). As expected, both asexual and sexual evolutionary synthesis produced networks with slight decreases in testing accuracy (approximately $3\%$); however, note that sexual evolutionary synthesis produced this network by generation 8 with a testing accuracy of $97.40\%$ while asexual evolutionary produced a corresponding network at generation 13 with a testing accuracy of $97.09\%$. 

Figure~\ref{fig_MNISTResults} (b) shows the synaptic sparsity for networks synthesized using asexual and sexual evolutionary synthesis and evaluated using the MNIST dataset. While networks synthesized via sexual evolutionary synthesis and asexual evolutionary synthesis produced similar testing accuracies, the network synthesized using sexual evolutionary synthesis at generation 8 has a synaptic efficiency of 258.37$\times$ and the network synthesized using asexual evolutionary synthesis at generation 13 has a synaptic efficiency of 139.37$\times$. Figure~\ref{fig_MNISTResults} (b) shows that the synaptic efficiency of networks synthesized using sexual evolutionary synthesis increases more steeply over generations than the synaptic efficiency of networks synthesized using asexual evolutionary synthesis. With almost double the synaptic efficiency for comparable testing accuracy, this indicates that networks synthesized via sexual evolutionary synthesis produce notably more efficient and compact feature representations.

Figure~\ref{fig_MNISTResults} (c) shows the cluster sparsity for networks synthesized using asexual and sexual evolutionary synthesis and evaluated using the MNIST dataset. Like synaptic efficiency, the cluster efficiency of networks synthesized using sexual evolutionary synthesis is noticeably higher than the cluster efficiency of networks synthesized using asexual evolutionary synthesis, and Figure~\ref{fig_MNISTResults} (c) shows a similar trend of increasing cluster efficiency over generations for networks synthesized via sexual evolutionary synthesis relative to asexual evolutionary synthesis. Specifically, the cluster efficiency of the network synthesized using sexual evolutionary synthesis at generation 8 is 34.29$\times$ while the cluster efficiency of the network synthesized using asexual evolutionary synthesis at generation 13 is 14.12$\times$, further indicating the potential of sexual evolutionary synthesis to produce more efficient and compact feature representations.

Figure~\ref{fig_CIFARResults} shows the CIFAR-10 testing accuracy and network efficiency (synaptic and cluster) for networks synthesized using asexual (red) and sexual (blue) evolutionary synthesis. Figure~\ref{fig_CIFARResults} (a) shows the testing accuracy for networks synthesized using asexual and sexual evolutionary synthesis. In this experiment, the original fully-trained ancestor network (generation 1) had a testing accuracy of $77.09\%$ with 144,736 synapses and 7,264 kernels (corresponding to a 3-channel input LeNet architecture~\cite{LeCun1998_LeNet}). While both asexual and sexual evolutionary synthesis produced networks with decreased testing accuracy (approximately $3\% - 4\%$), sexual evolutionary synthesis produced this network by generation 4 with a testing accuracy of $74.05\%$ while asexual evolutionary produced a corresponding network at generation 7 with a testing accuracy of $73.36\%$. Similarly for a testing accuracy decrease of approximately $10\%$, sexual evolutionary synthesis produced this network by generation 5 with a testing accuracy of $67.92\%$ while asexual evolutionary produced a corresponding network at generation 10 with a testing accuracy of $67.87\%$. 

Figure~\ref{fig_CIFARResults} (b) shows the synaptic sparsity for networks synthesized using asexual and sexual evolutionary synthesis and evaluated using the CIFAR-10 dataset. With the $3\% - 4\%$ drop in testing accuracy, the networks synthesized via sexual and asexual evolutionary synthesis have similar synaptic efficiencies of 10.66$\times$ and 13.76$\times$, respectively. At the $10\%$ testing accuracy drop, the network synthesized using sexual evolutionary synthesis at generation 5 has a synaptic efficiency of 36.27$\times$ and the network synthesized using asexual evolutionary synthesis at generation 10 has a synaptic efficiency of 30.66$\times$. While only a slight increase, this increase in synaptic efficiency still allows for somewhat more compact feature representations, and networks synthesized using sexual evolutionary synthesis exhibit a more exponential increase in synaptic efficiency relative to asexual evolutionary synthesis.

Figure~\ref{fig_CIFARResults} (c) shows the cluster sparsity for networks synthesized using asexual and sexual evolutionary synthesis and evaluated using the CIFAR-10 dataset. Unlike the synaptic efficiency where the increase at the $10\%$ testing accuracy drop is marginal, however, the cluster efficiency of networks synthesized using sexual evolutionary synthesis is notably higher than the cluster efficiency of networks synthesized using asexual evolutionary synthesis. Quantitatively, there is marginal improvement in cluster efficiency when comparing the networks with $3\% - 4\%$ drop in testing accuracy; the cluster efficiency of the network synthesized using sexual evolutionary synthesis is 2.20$\times$ at generation 4, and the cluster efficiency of the network synthesized using asexual evolutionary synthesis is 1.96$\times$ at generation 7. At the $10\%$ testing accuracy drop, the difference in cluster efficiency becomes more pronounced. The cluster efficiency of the network synthesized using sexual evolutionary synthesis at generation 5 is 4.82$\times$ while the cluster efficiency of the network synthesized using asexual evolutionary synthesis at generation 10 is 3.06$\times$. A more obvious increase relative to synaptic efficiency both in terms of increase in cluster efficiency over generations and quantitatively, the increase in cluster efficiency of networks synthesized via sexual evolutionary synthesis allows for more compact networks and feature representations.

Notice that in both the experiments with MNIST and CIFAR-10 datasets, fewer generations were required to reach similar levels of network performance (in this study, the expected drop in testing accuracy) using sexual evolutionary synthesis. It is worth noting that achieving a more efficient deep neural network at earlier generations is beneficial as it reduces the number of training steps required, which is the most computationally complex aspect of the evolutionary process. While more obvious in the experiment with the MNIST dataset, it can also be seen that sexual evolutionary synthesis produced networks that are more efficient on both the synaptic and cluster levels. This is particularly noticeable when comparing the trends of synaptic and cluster efficiency increase over generations, as the increases in efficiency for networks synthesized via sexual evolutionary synthesis follow a steeper exponential trend than networks synthesized via asexual evolutionary synthesis. Lastly, it is worth noting that the MNIST dataset allows for considerably more increases in architectural efficiency relative to the CIFAR-10 dataset; this is likely due to the simplicity of the MNIST dataset (1-channel images of handwritten digits) relative to the CIFAR-10 dataset (3-channel natural images of objects).
%More interestingly, however, the rates of this trade-off differ between the two sections of Table~\ref{tab_Results_CIFAR10}. Though both the asexual and sexual reproduction methods result in offspring networks with testing accuracy of approximately $74\%$ at the $11^{\text{th}}$ generation, it can be observed that the offspring network synthesized via asexual reproduction has an architectural efficiency of only $4.81$X while the offspring network synthesized via sexual reproduction has almost double the architectural efficiency at $9.12$X. This implies that the combination of two parent networks has the potential to significantly improve the overall adaptability of synthesized neural networks in the subsequent generations.

%% file: content/Conclusion.tex
In this work, we explored the effects of sexual evolutionary synthesis when synthesizing offspring deep neural networks in an evolutionary deep intelligence approach. An extension of the cluster-driven genetic encoding scheme proposed by Shafiee \textit{et al.}~\cite{Shafiee2016_2}, we incorporated a second parent network into the offspring network synthesis process at each generation.

Overall, the use of sexual evolutionary synthesis showed noticeable improvement in the architectural efficiency of the synthesized networks with maintaining comparable testing accuracy. In the case of the MNIST dataset, the offspring network synthesized via asexual evolutionary synthesis at generation 13 had a cluster efficiency of only 14.12$\times$ and a synaptic efficiency of 139.37$\times$ while the offspring network synthesized via sexual evolutionary synthesis at generation 8 had approximately double the architectural efficiency (cluster efficiency of 34.29$\times$ and synaptic efficiency of 258.37$\times$); both networks had a testing accuracy of around $97\%$. Similarly for the CIFAR-10 dataset, the offspring network synthesized via asexual evolutionary at generation 10 had a cluster efficiency of only 3.06$\times$ and a synaptic efficiency of 30.66$\times$ while the offspring network synthesized via sexual evolutionary synthesis at generation 5 had a cluster efficiency of 4.82$\times$ and a synaptic efficiency of 36.27$\times$; both networks had a testing accuracy of around $68\%$.

This suggests that sexual evolutionary synthesis in evolutionary deep intelligence via the synthesis of offspring neural networks using two parent networks can produce more efficient network architectures and increasingly compact feature representations, allow for higher levels of generalizability and adaptability in synthesized networks, and encourage more diversity in the genetic encoding. As such, further investigation into the effects of sexual versus asexual evolutionary synthesis for network synthesis in evolutionary deep intelligence would be beneficial, particularly with the deep neural networks in a changing environment, i.e., non-stationary environmental factor models.

Future work in this area includes a more thorough investigation into various methods for combining the parent neural networks, e.g., determining whether favouring the genetics of the parent with a better fitness score (such as testing accuracy) would allow for a higher probability of passing on beneficial traits to the offspring network and subsequent generations. Other potential areas of future work include associating strong task performance with specific synapses or synaptic clusters in a deep neural network, and favouring these specific sections of different parent networks during the offspring synthesis process. Lastly, it would be beneficial to extend the mating and synthesis process to include mate selection from a pool of potential neural network mates to incorporate the notion of ``survival of the fittest''~\cite{Williams2008} into the evolutionary deep intelligence approach to allow for a stronger overall population of networks.